%% file: main.tex
\documentclass[10pt,twocolumn,letterpaper]{article}

\usepackage[pagenumbers]{main}

\usepackage[table]{xcolor}
\usepackage{microtype}
\usepackage{graphicx}
\usepackage{booktabs}
\usepackage{amsmath}
\usepackage{amssymb}
\usepackage{mathtools}
\usepackage{amsthm}
\usepackage{color}
\usepackage{enumitem}
\usepackage{fontawesome5}
\usepackage{multirow}
\usepackage{makecell}
\usepackage{algorithmic}
\usepackage{algorithm}
\usepackage{etoolbox,siunitx}

\definecolor{lblue}{RGB}{231, 66, 52}
\usepackage[pagebackref,breaklinks,colorlinks,citecolor=lblue]{hyperref}
\def\paperID{1}
\def\confName{RoboSense}
\def\confYear{IROS 2025}

\title{A Parameter-Efficient Mixture-of-Experts Framework for Cross-Modal Geo-Localization}

\author{
    LinFeng Li\textsuperscript{1,2,*} \qquad
    Jian Zhao\textsuperscript{1,*,$\dagger$} \qquad
    Zepeng Yang\textsuperscript{1} \qquad
    Yuhang Song\textsuperscript{3} \qquad
    Bojun Lin\textsuperscript{3}
    \\ 
    Tianle Zhang\textsuperscript{1,$\dagger$} \qquad
    Yuchen Yuan\textsuperscript{1,$\dagger$} \qquad
    Chi Zhang\textsuperscript{1,$\dagger$} \qquad
    Xuelong Li\textsuperscript{1,$\dagger$}
    \\[1.2em] 
    \textsuperscript{1}The Institute of Artificial Intelligence (TeleAI), China Telecom \\
    \textsuperscript{2}East China Normal University \\
    \textsuperscript{3}National Tsing Hua University
}

\begin{document}

\maketitle
\let\thefootnote\relax\footnotetext{* These authors contributed equally to this work.}
\let\thefootnote\relax\footnotetext{$\dagger$ Corresponding authors.}
\input{sections/0_abstract}

\input{sections/1_intro}
\input{sections/2_related_work}
\input{sections/3_method}
\input{sections/4_experiments}

\input{sections/5_conclusion}

{
\small
\bibliographystyle{ieeenat_fullname}
\bibliography{main}
}

\end{document}

%% file: sections/0_abstract.tex
\begin{abstract}
We present a winning solution to RoboSense 2025 Track 4: Cross-Modal Drone Navigation. The task retrieves the most relevant geo-referenced image from a large multi-platform corpus (satellite/drone/ground) given a natural-language query. Two obstacles are severe inter-platform heterogeneity and a domain gap between generic training descriptions and platform-specific test queries. We mitigate these with a domain-aligned preprocessing pipeline and a Mixture-of-Experts (MoE) framework: (i) platform-wise partitioning, satellite augmentation, and removal of orientation words; (ii) an LLM-based caption refinement pipeline to align textual semantics with the distinct visual characteristics of each platform. Using BGE-M3 (text) and EVA-CLIP (image), we train three platform experts using a progressive two-stage, hard-negative mining strategy to enhance discriminative power, and fuse their scores at inference. The system tops the official leaderboard, demonstrating robust cross-modal geo-localization under heterogeneous viewpoints.
\end{abstract}

%% file: sections/1_intro.tex
\section{Introduction}
\label{sec:intro}

Cross-modal geo-localization, which aims to retrieve geo-referenced images from heterogeneous sources given natural language or visual queries, has emerged as a fundamental capability for autonomous navigation, situational awareness, and emergency response~\citep{bib01,bib02,bib03}. In particular, unmanned aerial vehicles (UAVs) play an increasingly critical role in tasks such as disaster management, infrastructure inspection, and urban planning, where robust geo-localization enables accurate scene understanding under diverse viewpoints~\citep{bib04,bib05}. However, building a generalizable model for cross-modal retrieval across drastically different platforms—satellite, drone, and ground-level imagery—remains highly challenging.

Two key obstacles hinder progress in this domain. First, the data heterogeneity across platforms introduces severe appearance gaps: satellite imagery exhibits large-scale, top-down structures, drone imagery captures mid-level oblique views, while ground-view images contain rich local details with clutter and occlusion. These discrepancies render a single, unified model less effective. Second, a significant domain gap exists between training and evaluation texts: training captions are often generic or verbose, whereas test queries are concise and intent-driven. More critically, the semantic focus of the descriptions often mismatches the visual modality (e.g., a generic caption may fail to capture the specific details relevant to a satellite or drone perspective), leading to poor generalization.

Existing approaches in vision-language retrieval typically rely on large pre-trained encoders such as CLIP~\citep{bib06,bib07,bib08,bib09,bib10} or ALIGN~\citep{bib11,bib12,bib13} to learn a shared embedding space. While effective on in-domain benchmarks, these methods often struggle to reconcile heterogeneous views and distributional discrepancies without costly fine-tuning on massive curated datasets. Ensemble strategies and Mixture-of-Experts (MoE)~\citep{jiang2024mixtral} methods offer a promising direction by combining specialized models, but most existing designs incur high parameter overhead or lack mechanisms to bridge textual domain gaps.

To address these challenges, we propose the Parameter-Efficient Mixture-of-Experts (PE-MoE) framework, a divide-and-conquer solution that integrates domain-aligned preprocessing with a lightweight expert design. Our framework partitions the dataset by platform, enabling each expert to specialize in satellite, drone, or ground imagery, while sharing a frozen backbone of strong pre-trained encoders (BGE-M3~\citep{bge-m3} for text, EVA-CLIP~\citep{sun2023eva} for images) to preserve generalization. To reduce the textual domain gap, we introduce an LLM-based caption refinement strategy. This process automatically revises captions to ensure their semantic focus aligns with the visual modality (e.g., emphasizing spatial relations for satellite images vs. object details for drone images), creating more precise training pairs. For satellite imagery, we further apply targeted augmentations alongside directional-text sanitization to ensure semantic consistency. The experts are trained using a progressive two-stage, hard-negative mining strategy to sharpen their discriminative abilities. Finally, a dynamic gating network adaptively routes queries to the most relevant experts, producing a fused similarity score.

This design achieves robust retrieval under severe viewpoint and modality shifts while maintaining parameter efficiency. On the RoboSense 2025 Track 4: Cross-Modal Drone Navigation, our method ranked first on the official leaderboard, demonstrating superior performance and strong generalization. Beyond competition success, our study highlights the importance of jointly addressing data heterogeneity and domain alignment, opening new directions for efficient cross-modal geo-localization.

%% file: sections/2_related_work.tex
\section{Related Work}
\label{sec:rea}
\subsection{Cross-Modal Image-Text Retrieval}
Cross-modal retrieval methods aim to learn a shared embedding space where images and texts can be aligned. Early approaches relied on recurrent encoders for text and CNN-based visual features optimized with triplet losses. With the advent of large-scale pre-training, methods such as CLIP~\cite{bib14,bib15,bib16,bib17,bib18}, ALIGN~\cite{bib19,bib20,bib21,bib22,bib23}, and BLIP~\cite{bib24,bib25,bib26} significantly advanced performance by leveraging large-scale image-text pairs. More recent work, e.g., BLIP-2~\cite{bib27,bib28,bib29,bib30,bib31}, explores parameter-efficient pre-training with frozen encoders and lightweight adapters. However, these models typically assume homogeneous data domains, and their performance degrades when facing drastic viewpoint shifts or domain gaps, as in UAV-based geo-localization.

\subsection{Visual Geo-Localization}
Visual geo-localization focuses on matching visual observations to geo-referenced imagery. Traditional methods include local feature matching\cite{bib32} and structure-based retrieval~\cite{bib33,bib34}, but they struggle with large viewpoint and scale changes. With deep learning, cross-view matching has gained momentum, particularly for ground-to-aerial matching tasks~\cite{bib25,bib26,bib27}. For instance, CVUSA~\cite{bib28} and University-1652~\cite{bib29} datasets highlight the importance of aligning satellite, drone, and ground perspectives. Despite progress, these methods remain challenged by domain heterogeneity and by the mismatch between verbose training captions and concise queries in real applications.

\subsection{Mixture-of-Experts and Model Ensembles}
Model ensembles and Mixture-of-Experts (MoE) approaches offer a promising way to enhance robustness by combining specialized models. Classical ensemble methods aggregate independent learners, while MoE frameworks introduce expert specialization with a gating network for adaptive routing~\cite{bib30,bib31,bib32}. Recent advances in parameter-efficient MoE integrate frozen backbones with lightweight expert modules, achieving strong trade-offs between specialization and scalability. In multimodal domains, MoE designs have been explored for vision-language pre-training~\cite{bib33,bib34,bib35}, but their application to UAV cross-modal geo-localization remains underexplored. Our work builds on this line by introducing a parameter-efficient MoE framework with domain-aligned preprocessing, enabling both specialization to platform-specific imagery and improved generalization across modality gaps.

%% file: sections/3_method.tex
\section{Method}

\begin{figure*}[h!]
    \centering
    \includegraphics[width=0.9\textwidth]{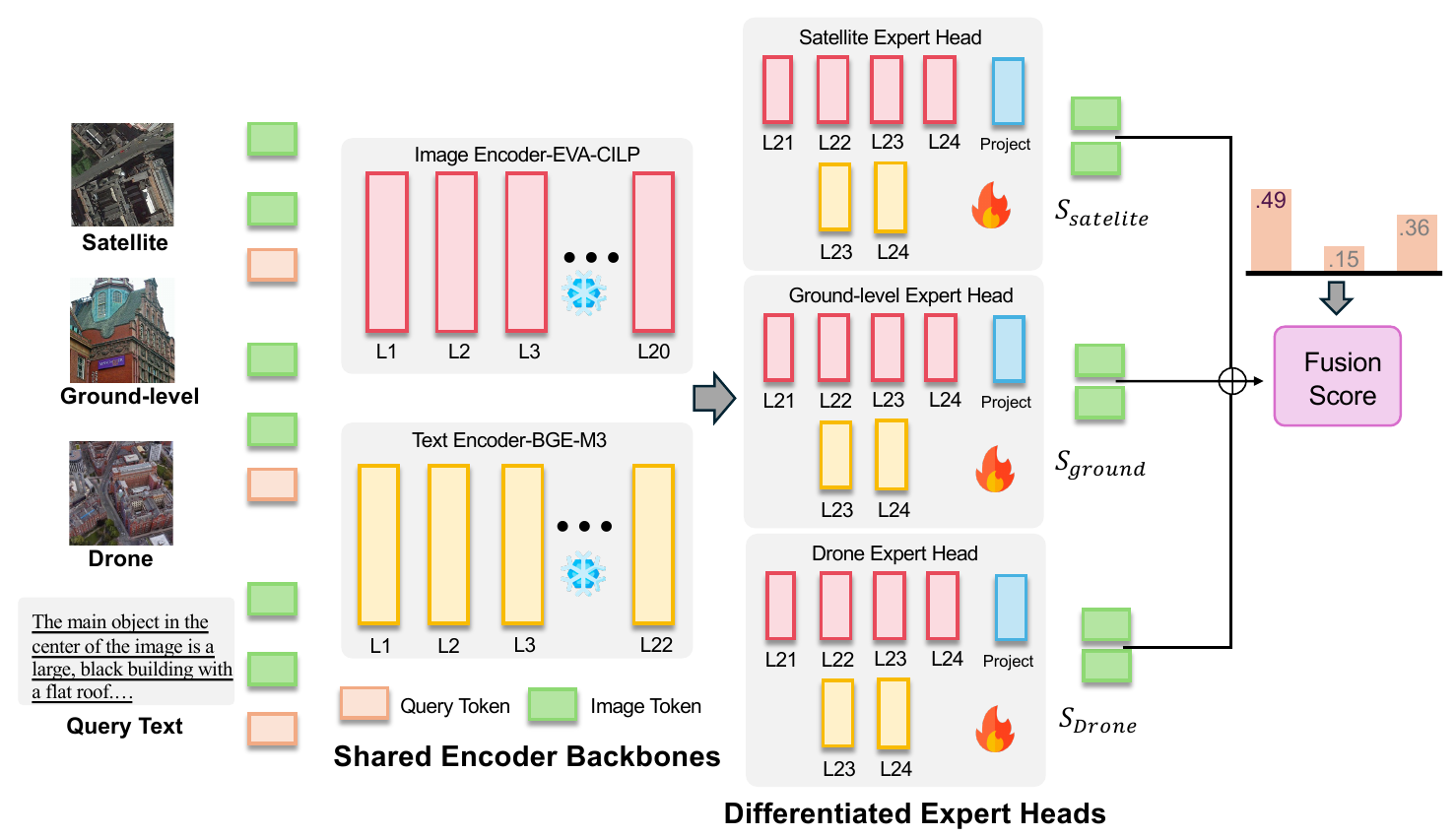} 
    \caption{The overall architecture of our Parameter-Efficient Mixture-of-Experts (PE-MoE) framework. A shared backbone extracts general features, which are processed by a dynamic gating network and specialized expert heads to produce the final retrieval score.}
    \label{fig:architecture}
\end{figure*}

In this chapter, we elaborate on the technical framework of our proposed solution, the Parameter-Efficient Mixture-of-Experts (PE-MoE). Our core philosophy follows a ``divide and conquer" principle, aiming to efficiently address the challenges of data heterogeneity and domain gaps by sharing generalized knowledge while specializing in specific domains. As illustrated in Figure~\ref{fig:architecture}, our framework is comprised of three primary stages: data preprocessing and alignment, the PE-MoE model architecture, and a two-stage training strategy.

\subsection{Data Preprocessing and Alignment}
We posit that targeted data preprocessing is a critical prerequisite for model success. Our strategy focuses on stratifying data by domain and aligning the textual distributions between training and testing phases.

\paragraph{Platform-based Data Stratification}
To tackle the profound visual discrepancies across platforms, we first partition the entire training dataset, $D$, into three distinct, non-overlapping subsets based on the image source: a satellite imagery subset, $D_{sat}$; a drone imagery subset, $D_{drone}$; and a ground-view imagery subset, $D_{ground}$. This stratification allows us to train highly specialized expert models for each visual domain.

\paragraph{Textual Domain Alignment}
We identified a significant domain gap in the textual descriptions relative to their corresponding image modalities. For example, the focus of a caption for a satellite image should differ substantially from that of a drone-view image (e.g., broad area relations vs. specific object details). To address this, we employed an LLM-based Caption Refinement strategy. We utilized a Large Language Model (LLM) to review and revise the caption for each training image. This process ensured that the textual description was semantically aligned with the image's specific visual perspective (satellite, drone, or ground). By tailoring the captions to be domain-specific, we provide the model with more accurate and consistent text-image pairs, enhancing the specialization of each expert.

\paragraph{Augmentation and Sanitization for Satellite Imagery}
Given the relatively small sample size of the satellite subset $D_{sat}$, we applied a series of data augmentation techniques, including random geometric transformations (e.g., rotations, flips) and photometric adjustments (e.g., brightness, contrast jitter). However, geometric transformations can alter the absolute spatial orientation of an image, creating semantic inconsistencies with textual descriptions containing directional language (e.g., "to the north of," "on the left side"). To resolve this, we employed a complementary text sanitization process. Before applying geometric augmentations, a keyword-matching algorithm automatically removed any sentences with explicit directional phrases from the corresponding captions, ensuring semantic consistency between the augmented images and their textual descriptions.

\subsection{Parameter-Efficient MoE Framework}
Our model architecture is designed to achieve maximum specialization with minimal parameter overhead.

\paragraph{Shared Encoder Backbones}
We utilize the state-of-the-art BGE-M3~\citep{bge-m3} as our text encoder and EVA-CLIP~\citep{EVA-CLIP} as our image encoder. To maximize parameter efficiency and preserve their powerful, general-purpose representational abilities, the vast majority of the parameters in these backbone models are \textbf{kept frozen} during training. Any input text or image undergoes a single forward pass through these shared backbones to yield high-level, generalized feature representations, denoted as $t_{\text{shared}}$ and $v_{\text{raw\_shared}}$.

\paragraph{Differentiated Expert Heads}
Building upon the shared backbones, we designed three lightweight expert heads, one for each platform: $H_{sat}, H_{drone}, \text{and } H_{ground}$. Each expert head is an independent, trainable module comprising:
\begin{itemize}
    \item The final few (e.g., 2) trainable transformer layers of the BGE-M3 and EVA-CLIP models.
    \item A distinct, trainable visual projection layer that maps image features into the common embedding space.
\end{itemize}
Each expert head $H_k$ is trained exclusively on its corresponding data subset $D_k$. It takes the shared features as input and processes them to generate domain-specific final embeddings $(t_k, v_k)$, from which a similarity score $S_k(q, I) = \text{cosine}(t_k, v_k)$ is computed.

\paragraph{Dynamic Gating Network}
To intelligently orchestrate the experts, we designed a dynamic gating network, $G$. It is a small, two-layer Multi-Layer Perceptron (MLP) that takes the shared text feature $t_{\text{shared}}$ as input. Its output is a 3-dimensional logits vector, which is passed through a Softmax function to produce a query-dependent weight distribution $g(q) = [g_{sat}, g_{drone}, g_{ground}]$, where $\sum_k g_k(q) = 1$. The gate learns to "understand" the query's intent and assign the highest weight to the expert best suited to handle it.

\subsection{Training and Inference}

\begin{figure}[h!] 
    \centering
    \includegraphics[width=0.8\columnwidth]{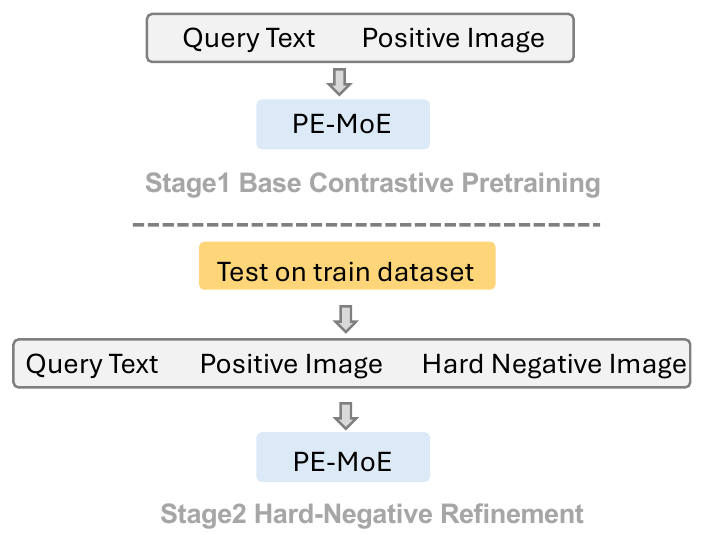} 
    \caption{The two-stage training pipeline. Stage 1 builds general alignment using positive pairs, while Stage 2 uses mined hard negatives to refine the model's discriminative ability.}
    \label{fig:training_pipeline}
\end{figure}

\paragraph{Two-Stage Training Strategy}
As illustrated in Figure~\ref{fig:training_pipeline}, our training follows a progressive two-stage strategy.

In Stage 1 (Base Contrastive Pretraining), we train the PE-MoE model on positive text–image pairs using contrastive learning. This stage aims to build a robust general alignment between textual and visual representations across the diverse domains.

Following this, we perform an intermediate step where we test the model on the training set itself. This process allows us to efficiently mine hard negative samples (i.e., images that are semantically incorrect but have high similarity scores) for each query.

In Stage 2 (Hard-Negative Refinement), we retrain the model, this time providing it with triplets of (query text, positive image, hard negative image). This stage sharpens the model's discriminative ability, forcing it to learn the subtle differences between correct and highly similar incorrect images. This progressive strategy significantly improves model robustness under heterogeneous domains without increasing the total parameter count.

\paragraph{Inference Process}
During inference, for a given text query $q$ and a candidate image $I$, the final similarity score is computed as a dynamically weighted sum of the individual expert scores. The entire process is formalized in Equation~\ref{eq:final_score}.
\begin{equation}
\label{eq:final_score}
S_{\text{final}}(q, I) = \sum_{k \in \{\text{sat, drone, ground}\}} g_k(q) \cdot S_k(q, I)
\end{equation}
All candidate images in the gallery are ranked based on this final score $S_{\text{final}}$ to produce the retrieval results.

%% file: sections/4_experiments.tex
\section{Experiments}
\label{chap:experiments}

This chapter presents a series of experiments designed to validate the efficacy of our proposed PE-MoE framework. We detail our experimental setup, present our main results in the competition, and conduct in-depth ablation studies to analyze the contribution of each component.

\subsection{Experimental Setup}
\paragraph{Dataset} All experiments were conducted on the official dataset for the RoboSense 2025 Track 4 challenge, University-1652. We strictly adhered to the official data splits and task definition for text-to-image retrieval.

\paragraph{Evaluation Metrics} We adopted the official evaluation metrics for the challenge, which are Recall at K (R@K) for K=1, 5, and 10. R@K measures the percentage of queries for which the correct gallery image is retrieved within the top K results.

\subsection{Implementation Details}
Our framework was implemented in PyTorch. The shared backbones were initialized from the pre-trained weights of \texttt{bge-m3-base} and \texttt{eva-clip-large}. Each expert head consisted of the final two trainable transformer layers of text encoder, the final four trainable transformer layers of image encoder and a linear projection layer to map visual features to a 1024-dimensional space. The gating network was a 2-layer MLP with a 512-dimensional hidden layer. We used the AdamW optimizer with a learning rate of $2 \times 10^{-5}$ and a weight decay of $1 \times 10^{-4}$. All images were resized to $384 \times 384$ pixels. The models were trained on eight NVIDIA A100 (80GB) GPUs with a batch size of 128.

\subsection{Main Results}
Our proposed PE-MoE framework achieved state-of-the-art performance on the official test set, securing first place on the final leaderboard. Table~\ref{tab:main_results} presents a comparison of our results against the official baseline and other top-performing teams. The results clearly demonstrate the superiority of our approach across all key metrics.

\definecolor{highlightcolor}{HTML}{E1F5FE}

\begin{table}[h!]
\centering
\caption{Performance comparison on the University-1652 test set leaderboard.}
\label{tab:main_results}
\begin{tabular*}{\columnwidth}{@{\extracolsep{\fill}}lcccc@{}} 
\toprule
\textbf{Method} & \textbf{R@1} & \textbf{R@5} & \textbf{R@10}  & \textbf{Score}\\ \midrule
Official Baseline & {25.44} & {40.61} & {49.10} & {39.27} \\
2nd Place & {28.34} & {54.08} & {66.11} & {47.23} \\
3rd Place & {31.33} & {49.09} & {57.15} & {44.24} \\ \midrule
\rowcolor{highlightcolor}
\textbf{Our PE-MoE} & \textbf{38.31} & \textbf{53.70} & \textbf{61.32} & \textbf{49.82} \\ \bottomrule
\end{tabular*}
\end{table}

\begin{table*}[h!]
\centering
\caption{Ablation analysis of the components in our proposed framework.}
\label{tab:ablation_study}
\begin{tabular}{@{}clcccc@{}}
\toprule
\textbf{\#} & \textbf{Model Configuration} & \textbf{R@1} & \textbf{R@5} & \textbf{R@10} & \textbf{Score} \\ \midrule
1 & Baseline: Unified Model w/o Preprocessing & {21.32} & {35.90}  & {42.01} & {31.67} \\
2 & + Textual Domain Alignment & {27.87} & {45.13} & {53.22} & {40.55} \\
3 & + Static Ensemble of Expert Heads & {34.42} & {49.77} & {58.23} & {46.33} \\
\rowcolor{highlightcolor}
4 & \textbf{Full Model: PE-MoE w/ Dynamic Gating} & \textbf{38.31} & \textbf{53.70} & \textbf{61.32} & \textbf{49.82}  \\ \bottomrule
\end{tabular}
\end{table*}

\subsection{Ablation Studies}
To rigorously evaluate the contribution of each component in our framework, we conducted a comprehensive ablation study. We started with a basic unified model and progressively added our proposed techniques. The results are summarized in Table~\ref{tab:ablation_study}.

\paragraph{Analysis} The results from our ablation study lead to several key insights. First, comparing model \#2 to \#1, the introduction of our textual domain alignment strategy yields a significant improvement in R@1, confirming its crucial role in mitigating the text domain gap. Second, the transition from model \#2 to \#3, which replaces the unified model with specialized expert heads (fused with static weights), results in another substantial performance leap. This validates our core "divide and conquer" hypothesis. Finally, comparing our full model (\#4) to the static ensemble (\#3), the dynamic gating network provides a further discernible boost in accuracy. This demonstrates that an intelligent, query-aware routing mechanism is superior to a fixed-weight fusion, allowing the system to adaptively leverage the best expert for each specific query. Together, these components synergistically contribute to the overall state-of-the-art performance of our final model.

%% file: sections/5_conclusion.tex
\section{Conclusion}
In this work, we presented a winning solution to RoboSense 2025 Track 4: Cross-Modal Drone Navigation. To address the challenges of severe platform heterogeneity and textual domain gaps, we proposed a Parameter-Efficient Mixture-of-Experts (PE-MoE) framework combined with a domain-aligned preprocessing pipeline. Specifically, our approach partitions data by platform, augments scarce satellite imagery while sanitizing captions, and aligns the training text distributions via sentence-level splitting. Built upon frozen pre-trained encoders (BGE-M3 and EVA-CLIP), lightweight expert heads specialize in distinct platforms, and a dynamic gating network adaptively routes queries for optimal retrieval.
Extensive experiments on the official benchmark demonstrated that our framework achieves state-of-the-art performance and ranked first on the leaderboard, validating its robustness and effectiveness in heterogeneous cross-modal geo-localization.
Looking forward, future research may focus on developing end-to-end trainable MoE frameworks, exploring dynamic routing strategies beyond simple softmax gating, and integrating multi-scale and temporal cues for enhanced UAV navigation in complex, real-world environments.